\documentclass[letterpaper]{article}
\usepackage{amsmath,epsfig}
\usepackage[preprint]{spconfa4}
\usepackage{microtype}
\usepackage{booktabs}
\usepackage{tabularx}
\usepackage[usestackEOL]{stackengine}
\usepackage{nicefrac}

\let\OLDthebibliography\thebibliography
\renewcommand\thebibliography[1]{
  \OLDthebibliography{#1}
  \setlength{\parskip}{0pt}
  \setlength{\itemsep}{0pt plus 0.3ex}
}

\begin{document}\sloppy

\def\x{{\mathbf x}}
\def\L{{\cal L}}

\title{Investigating Class-level Difficulty Factors in Multi-label Classification Problems}

\makeatletter
\def\@name{\textit{Mark Marsden}$^\ast$, \textit{Kevin McGuinness}$^\ast$, \textit{Joseph Antony}$^\ast$, \textit{Haolin Wei}$^\ast$, \textit{Milan Red\v{z}i\'c}$^\dagger$ \\* \textit{Jian Tang}$^\dagger$, \textit{Zhilan Hu}$^\ddagger$, \textit{Alan Smeaton}$^\ast$, \textit{Noel E. O'Connor}$^\ast$ \\}
\makeatother

\address{$^\ast$Insight SFI Research Centre for Data Analytics, Dublin City University, Ireland \\
$^\dagger$Huawei Ireland Research Center, Dublin, Ireland, \\
$^\ddagger$Huawei Technologies Co. Ltd, China}

\maketitle

\begin{abstract}
This work investigates the use of class-level difficulty factors in multi-label classification problems for the first time. Four class-level difficulty factors are proposed: frequency, visual variation, semantic abstraction, and class co-occurrence. Once computed for a given multi-label classification dataset, these difficulty factors are shown to have several potential applications including the prediction of class-level performance across datasets and the improvement of predictive performance through difficulty weighted optimisation. Significant improvements to mAP and AUC performance are observed for two challenging multi-label datasets (WWW Crowd and Visual Genome) with the inclusion of difficulty weighted optimisation. The proposed technique does not require any additional computational complexity during training or inference and can be extended over time with inclusion of other class-level difficulty factors.
\end{abstract}



\section{Introduction}
\label{sec:intro}
The concept of difficulty estimation has been investigated for several computer vision tasks including weakly-labeled object localisation \cite{tudor2016hard} and image classification \cite{scheidegger2018efficient}. This  approach assigns a difficulty score for the task to a given image or collection of images based on their content. This difficulty score quantifies how challenging it is to perform a given analysis task (e.g. retrieval, classification, etc.). Once computed,  difficulty scores can be used to predict the performance and to modify the training strategy of a predictive model to improve performance \cite{tudor2016hard}. To date, this approach has yet to be applied to {\em multi-label} image classification problems, where several classes can occur simultaneously for a given image or video. In this context a difficulty score is assigned to each class individually, indicating how challenging it can be to recognise a particular class in a multi-label classification problem.

\begin{figure}[t]
\centering
\includegraphics[width=0.42\textwidth, height=0.2\textwidth]{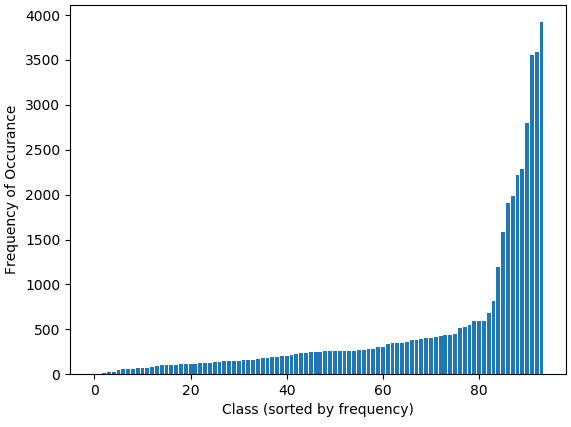}
\includegraphics[width=0.42\textwidth, height=0.2\textwidth]{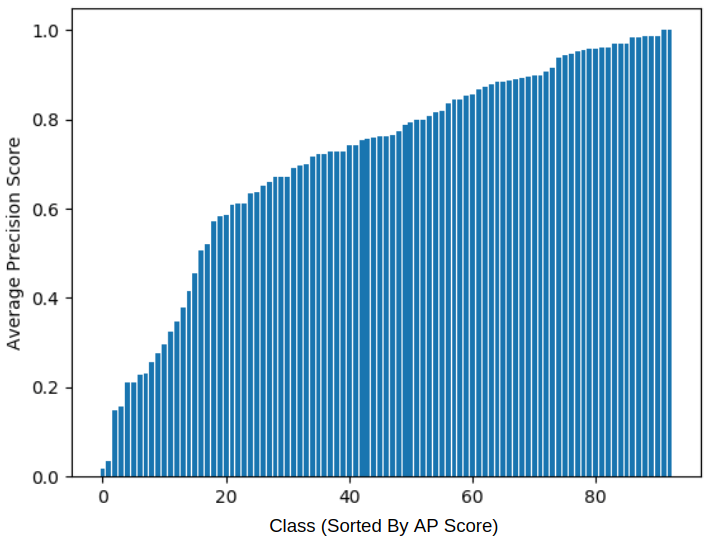}
\caption{Class frequency distribution in the WWW Crowd training set and range of AP performance scores observed on the WWW Crowd validation set \cite{shao2015deeply} using a Resnet50 baseline.}
\label{WWW_distribution}
\end{figure}

One of the main factors attributed to class-level performance in recognition tasks is the frequency with which a given class occurs within the available training data \cite{zhang2018binary}. One may see what impact class frequency has on performance in Figure~\ref{WWW_distribution}, which represents the distribution of class frequency within the WWW Crowd training set \cite{shao2015deeply}, and the range of class-level Average Precision (AP) scores achieved on the WWW crowd validation set using a Resnet50 CNN baseline approach \cite{he2016deep}. Significant variation in both class frequency and validation AP can be observed for this dataset, but the question remains as to how strongly linked these two factors are.

\begin{figure}[t]
\centering
\includegraphics[width=0.42\textwidth, height=0.25\textwidth]{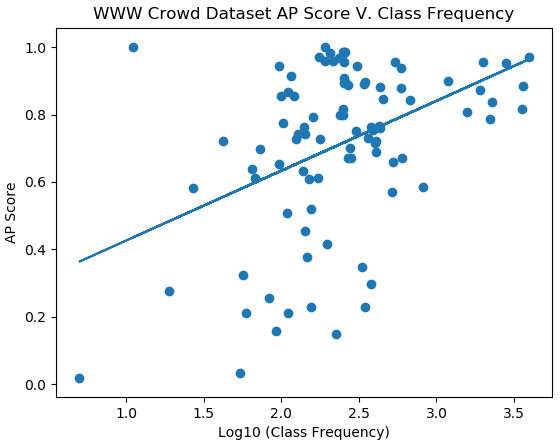}
\caption{Relationship between class frequency and AP validation performance on the WWW Crowd dataset \cite{shao2015deeply}.}
\label{WWW_freq_perf}
\end{figure}

Figure~\ref{WWW_freq_perf} shows the relationship between class frequency and AP validation performance on the WWW Crowd dataset, with a Pearson's correlation coefficient of 0.42. While there is a strong positive correlation between these variables, there are many outliers that do not follow this trend. Clearly, there are other confounding difficulty factors that influence class-level recognition performance in multi-label problems. Identifying and quantifying these confounding factors could potentially allow class-level performance to be predicted before training and to subsequently inform training strategies.

This paper investigates several class-level properties in addition to class frequency, as potential difficulty factors in multi-label classification tasks. These include the level of visual variation within the positive training samples for a given class, the level of semantic abstraction associated with a given concept keyword, and the level of co-occurrence with other concepts in the training set. Once measured, these difficulty factors can be combined to produce an overall difficulty score for each class. This score can be used for performance prediction, difficulty-informed training, and evaluation. This type of approach also allows for additional difficulty factors to be identified and included over time to refine the difficulty estimation.

This paper also looks to employ a more comprehensive evaluation strategy for multi-label classification in order to analyse the overall impact of model design choices on class-level performance. Our focus is on investigating general relationship between class-level properties and multi-label classification. The typical evaluation procedure for multi-label classification involves calculating a given performance metric, such as a ROC curve, AUC or Average Precision, for all classes individually before taking an arithmetic mean of class scores to measure overall recognition performance. This type of approach can often obfuscate poor and inconsistent performance between classes, producing a shallow and misleading impression of overall system performance. In order to address this issue, class-level performance box plots are combined with the traditional arithmetic mean to provide a more comprehensive and informative performance evaluation. 

The core contributions of this paper are:
1) The analysis of four class-level difficulty factors in multi-label classification (frequency, visual variation, semantic abstraction, co-occurrence; 2) The estimation of class-level performance across datasets for multi-label classification using difficulty factors; 3) Applying a difficulty weighted loss function to boost the predictive performance of a multi-label classification model; 4) A state-of-the-art predictive performance on the WWW Crowd dataset. 

\section{Related Work}

\subsection{Difficulty in Computer Vision}
Bengio et al. looked at difficulty in machine learning in the context of so called \textit{Curriculum Learning} \cite{bengio2009curriculum}, in which a perceptron is trained on progressively more difficult samples over time, improving the overall convergence time and predictive performance of the model. The difficulty of a given sample is measured in this work using the decision margin of an already trained perceptron classifier \cite{bengio2009curriculum}. 

Image-level difficulty was measured prior to model training by Ionescu {\em et al.} as the human response time in a visual search task measured in milliseconds \cite{tudor2016hard}. Using a set of response times, a support vector regression model was trained for difficulty prediction with CNN features and used to improve performance in weakly supervised object localization and classification tasks \cite{tudor2016hard}. The drawback of this approach is the cost in time and money for large scale human annotation.

Dataset-level difficulty was estimated by Scheidegger {\em et al.} for single-label classification tasks \cite{scheidegger2018efficient}. Dataset difficulty in this work corresponds to how easily separated the classes are in a collection of images. This is calculated using a probe network, which accurately predicts difficulty for a classification task 27$\times$ faster than training a full classification network~\cite{scheidegger2018efficient}.

\subsection{Multi-Label Classification Methods}
Binary relevance \cite{zhang2018binary} is a common approach to multi-label classification where the problem is decomposed into a set of $N$ binary classification tasks, where $N$ is the number of classes. This approach typically fails to exploit inter-class correlations, which can be used to help improve generalization \cite{zhang2018binary}. To address this issue the classifier chain method was developed, which ensures class correlation through a chain of binary classifiers while maintaining the computational complexity of conventional binary relevance~\cite{read2009classifier}.

Zhang {\em et al.} applied neural networks to the multi-label classification problem through the development of the BP-MLL algorithm (backpropagation for multi-label learning) \cite{zhang2006multilabel}. In this approach a pairwise ranking error and hyperbolic tan activation are employed to allow neural networks to be trained to natively perform multi-label classification \cite{zhang2006multilabel}. The work of Zhang {\em et al.} was subsequently expanded upon by Nam {\em et al.} \cite{nam2014large} who included a cross-entropy based loss function and sigmoid activation, improving both convergence speed and overall performance. This approach combined with the end-to-end learning potential of the latest deep CNN architectures \cite{he2016deep} has lead to neural networks becoming the dominant approach for multi-label image classification \cite{shao2015deeply,wang2016cnn,zhu2015multi}.

Wang {\em et al.} tackled multi-label classification by employing ranking and thresholding heuristics to refine the set of predictions \cite{wang2016cnn}. In this work a joint image-label embedding is produced to capture semantic label dependency as well as image-label relevance \cite{wang2016cnn}. RethinkNet \cite{yang2018deep} reformulates the sequence prediction problem in the multi-label classification paradigm by adopting a recurrent neural network (RNN) to gradually refine  predictions and to store the label correlation information together with a subsequent class-level analysis. Ge {\em et al.} propose a weakly supervised curriculum learning pipeline for multi-label object recognition, detection, and semantic segmentation \cite{ge2018multi}. It employs an algorithm for filtering, fusing, and multi-label classifying object instances using class occurrences collected from multiple mechanisms~\cite{ge2018multi}. 


\section{Baseline Approach}

The baseline multi-label classification method employed in this work consists of a 50-layer Resnet CNN \cite{he2016deep} adapted to a multi-label setting. The number of neurons in the final fully connected layer is updated to reflect the required number of classes for a given dataset before a sigmoid activation is applied on the final output. We optimize the total binary cross entropy between the $K$ binary target attributes $S_j$ and the predictions $\hat{S_{j}}$:
\[
L_{\text{BCE}}=-\sum_{j=1}^{K}S_{j}\log(\hat{S_{j}})+(1-S_{j})\log(1-\hat{S_{j}}),
\]
using stochastic gradient descent (SGD) with a fixed learning rate of 0.001, a momentum of 0.9, and a batch size of 28 (The maximum batch size possible with the hardware used). Training is carried out for 100,000 iterations in all experiments, with the number of training set epochs varying based on size of the dataset in use. The network is initialised using weights from a 50-layer ResNet model pre-trained on ImageNet~\cite{imagenet_cvpr09}.

Aggressive training set augmentation is applied dynamically during optimisation, with random crops (dynamic size and aspect ratio) taken, random horizontal rotations applied as well as random alterations to brightness, contrast, saturation, and hue (colour jitter transformations). Normalisation is applied during training and inference using the channel-level means and standard deviations of the ImageNet collection \cite{imagenet_cvpr09}. 10-crop augmentation (corners, centre and horizontal flips) is applied for each sample at inference. For video datasets, the proposed single-frame model is trained using randomly selected key frames and the video-level labels. Inference on video datasets is performed by individually processing 10 equally-spaced keyframes before taking the mean prediction for the overall video clip.


\section{Datasets}
\subsection{WWW Crowd}
The WWW (Who What Why) Crowd dataset contains 10,000 clips of crowded scenes labelled for 94 crowd behavior concepts and related scene content labels \cite{shao2015deeply}. The average clip length is 23 seconds with a standard deviation of 26. This multi-label collection is divided into training, validation and test sets using a 7:1:2 split. Classification is carried out at the video level while evaluation is performed using mean Average Precision (MAP) and receiver operator characteristic AUC (area under the curve).  Examples of the class labels in this dataset include `walk,' `outdoor,' `skate,' `fight,' and `run.'  

\subsection{Visual Genome}
The Visual Genome dataset \cite{krishna2017visual} is a varied collection of 108,000 images densely annotated for attribute recognition, object classification and visual question answering (VQA). These images contain a variety of everyday scenes containing several distinct objects. 
Here the provided attribute annotations will be used to develop a challenging multi-label classification task. This collection contains 40,513 unique attribute tags; however, there is a lot of redundancy within this set of labels. The overall set of attribute labels is first pre-processed with whitespace removed, lowercase applied and stemming performed using WordNet~\cite{miller1995wordnet}. From the resulting set of attribute labels, the top 250 most frequently occurring attributes are used as the classes for the proposed multi-label classification task. Examples include `Plastic,' `Wooden,' `Round,' `Bright,' `Wet,' `Rocky,' and `Beige.' This dataset is also split into training, validation, and test sets using a 7:1:2 ratio. Performance is again evaluated using MAP and AUC.


\section{Measuring Class-Level Difficulty Factors}
Approaches to measure the proposed set of class-level difficulty factors for multi-label classification are defined in this section. Once calculated for a given dataset, any desired combination of these difficulty factors can be either combined to form a feature vector for performance prediction or summed to produce an overall difficulty score for each individual class.

\subsection{Class Frequency}
The class frequency factor, $Freq_{i}$ captures how often a given class $i$ occurs within a multi-label classification dataset. We define it to be:
\[
\mathrm{Freq}_{i}=\frac{ \log{s_{i}}}{\max\limits_{j} \log{s_{j}} },
\]
where $s_{i}$ is the  number of times class $i$ occurs in the dataset. The $\log$ transform is applied due to significant range in class frequency observed in many multi-label classification datasets (10s of samples vs. 1000s of samples). Division by the the maximum value is applied to scale this difficulty factor between 0.0 and 1.0.

\subsection{Visual Variation}
Visual variation corresponds to the variety in appearance of positive training samples for a given class. This difficulty factor is measured by first extracting high level CNN descriptors  from all training samples within a given dataset. An ImageNet pre-trained Resnet CNN \cite{he2016deep} is used to this end, with the output of the second to last layer used for feature extraction.  For each class $i$, the extracted features for all positive training samples are stacked together and a mean descriptor (centroid) calculated. The cosine distance between this centroid to each positive training sample for class $i$ is  then calculated and the maximum distance taken as the visual variation difficulty factor.  This distance is then divided by the maximum visual variation difficulty factor among the class datasets to scale the difficulty factor between 0.0 and 1.0.

\subsection{Semantic Abstraction}
The semantic abstraction difficulty factor corresponds to the degree to which the keyword associated with the class is  semantically abstract or concrete. This difficulty factor is calculated using a crowd sourced dataset obtained by Brysbaret {\em et al.} who carried out a study to produce concreteness scores for 40,000 English lemmas using 4,000 volunteers \cite{brysbaert2014concreteness}. A hypothesis of this work is that a more concrete label is easier to recognise in a multi-label problem than a highly abstract label. In this dataset each word has a concreteness rating between 0.0 and 5.0. In the event that the given keyword is not found in this dataset then the closest matching  text string is used. Once a concreteness score is found it is divided by 5.0 to scale this difficulty score between 0.0 and 1.0.

\subsection{Class Co-occurrence}
Class co-occurrence represents how often a given class occurs with other classes in a multi-label classification problem. This is defined for a given training set by first generating a ground truth matrix $Y$, where each row corresponds to a sample, each column corresponds to a class, 1.0 indicates the presence of the class and 0.0 the absence of it. The co-occurrence matrix is then calculated given by $C = Y^TY$. The diagonal of $C$ is then set to 0.0 (self co-occurrence is removed). 
The co-occurrence level for class $i$ can then be calculated as:
\[
\mathrm{Cooc}_{i}=\frac{\nicefrac{\sum\limits_{j} C_{ij}}{s_{i}}}{\max\limits_{k}\nicefrac{\sum\limits_{j} C_{kj}}{s_{k}}},
\]
where $s_{i}$ is total number of times the given class $i$ occurs in the training set. Division by $s_{i}$ normalises for class frequency.

\section{Difficulty Weighted Training}

Using the difficulty factors, it is possible to perform difficulty weighted training by applying a set of class-specific loss weights during CNN optimisation, producing a modified binary cross entropy loss. These class-specific loss weights are generated by computing a set of difficulty scores for the classes of a given dataset and taking the reciprocal for class score. The intention here is to apply a higher weighting for a more difficult class and a lower score for an easier class. The same CNN baseline approach is used as before, with just the computation of the loss function altered.

\subsection{Impact of Each Difficulty Factor}

\begin{figure*}[ht]
\centering
\includegraphics[width=0.4\textwidth, height=0.22\textwidth]{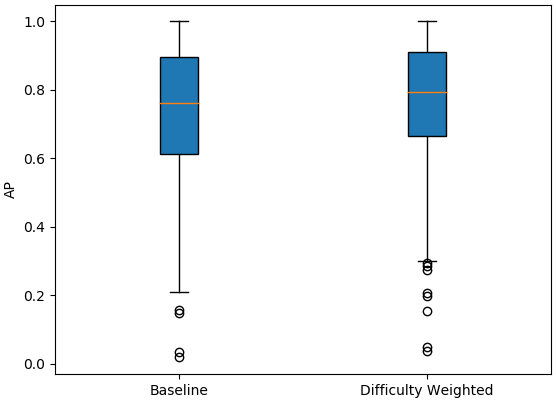}
\includegraphics[width=0.4\textwidth, height=0.22\textwidth]{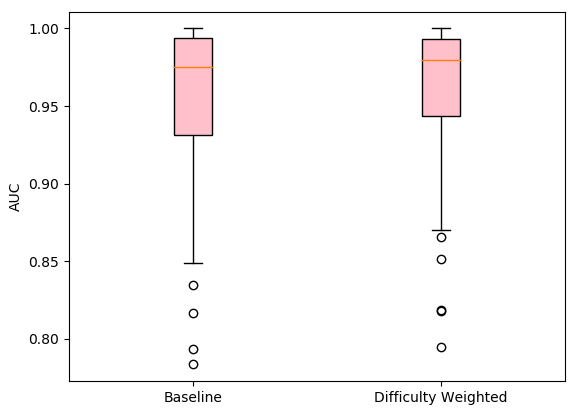}
\caption{Box plots highlighting the range of AP and AUC values observed for the WWW Crowd validation set using the baseline approach and when difficulty weighting is applied}
\label{boxplots}
\end{figure*}

Table~\ref{perf_weighted_WWW} presents the MAP and AUC on the WWW Crowd validation set when various difficulty factors are used individually as loss weights and also when they are combined. The best overall performance is achieved when all difficulty factors are combined, with a 4.3\% and 0.7\%  relative improvement over the baseline observed for MAP and AUC respectively. All difficulty factors are used when computing difficulty weights for all subsequent experiments. Figure~\ref{boxplots} presents a set of  box plots showing the range of class-level AP and AUC scores with and without difficulty weighting applied. Difficulty weighting is clearly shown to produce more consistent class-level performance centered around the median value.

\begin{table}[ht]
\centering
\begin{tabularx}{\columnwidth}{Xrr}
\toprule
\textbf{Weighting}            & \textbf{MAP}            & \textbf{AUC} \\ \midrule
None                 & 0.710           &  0.955   \\ 
Frequency            & 0.734          &  0.961  \\ 
Visual Variation     & 0.732          &  0.960   \\ 
Semantic Abstraction & 0.723          & 0.958    \\ 
Co-occurrence         & 0.729          &  0.959    \\ 
All Factors         & \textbf{0.743} &  \textbf{0.961}  \\ \bottomrule
\end{tabularx}

\caption{Validation performance on the WWW dataset with difficulty based loss weightings applied during training.}
\label{perf_weighted_WWW}
\end{table}

\subsection{Experiments}
The proposed difficulty weighted loss approach is compared to existing approaches in the literature for the WWW Crowd Dataset. Table~\ref{test_wwww} compares AUC and AP performance on the WWW Crowd test set with the state-of-the-art. The proposed approach achieves the best overall performance on the WWW crowd dataset, with a 2.7\% and 15.5\% relative improvement observed in AUC and MAP performance respectively. This significant improvement over  existing methods is achieved through a combination of difficulty weighted optimisation and a well implemented multi-label classification pipeline. 

\begin{table}[ht]
\centering
\begin{tabularx}{\columnwidth}{Xll}
\toprule
\textbf{Approach} & \textbf{MAP} & \textbf{AUC} \\ \midrule
DLSF+DLMF \cite{shao2015deeply}  & 0.412 & 0.877 \\ 
3D CNN \cite{ji20133d} & 0.39 & 0.86 \\ 
Slicing CNN \cite{shao2016slicing} & 0.6255 & 0.94 \\ 
Proposed & \textbf{0.723}  & \textbf{0.966} \\ \bottomrule
\end{tabularx}
\caption{Test set performance on the WWW crowd dataset.}
\label{test_wwww}
\end{table}

\begin{figure*}
\centering

\includegraphics[width=0.42\textwidth, height=0.24\textwidth]{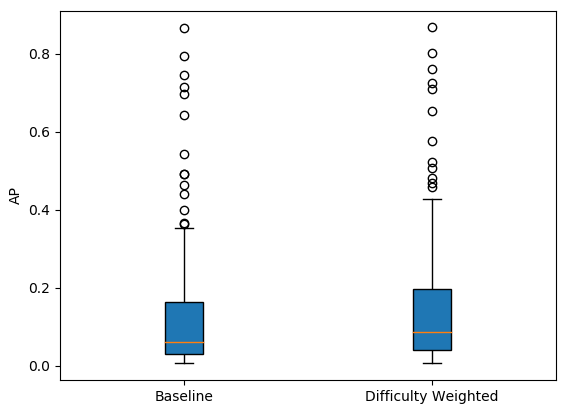}
\includegraphics[width=0.42\textwidth, height=0.24\textwidth]{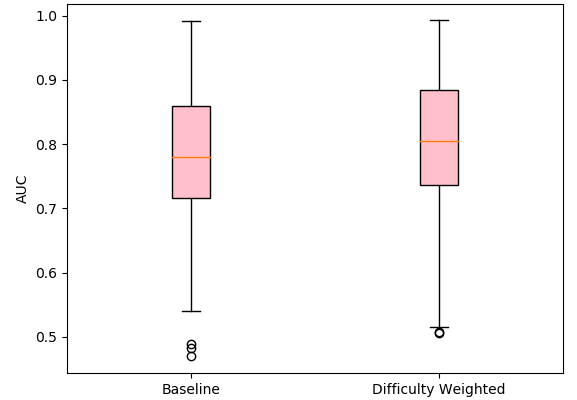}

\caption{Box plots highlighting the range of AP and AUC values observed for the Visual Genome test set using the baseline approach and when difficulty weighting is applied}
\label{boxplots_vg}
\end{figure*}

Test set performance on the Visual Genome dataset with and without difficulty weighted training is presented in Table~\ref{vg_test}. Again the application of difficulty weighted training results in improvements to both MAP and AUC performance, with a relative improvement of 15.3\% and 2.5\% for each metric. These improvements are similar to those observed for the WWW crowd dataset. Figure~\ref{boxplots_vg} presents box plots showing the range of class-level AP and AUC scores with and without difficulty weighting for the Visual Genome test set. Difficulty weighting is shown to again improve the overall class-level performance for both metrics.

The benefits of this approach are demonstrated for multiple datasets (WWW Crowd and Visual Genome), showing how difficulty weighted training can improve performance of any multi-label classification problem, requiring no increase in computational complexity during training or inference.

\begin{table}
\centering
\begin{tabularx}{\columnwidth}{Xrr}
\toprule
\textbf{Approach} & \textbf{MAP} & \textbf{AUC} \\ \midrule
Baseline & 0.121 & 0.781 \\ 
Difficulty Weighted & \textbf{0.141} & \textbf{0.801} \\ \bottomrule
\end{tabularx}
\caption{Test set performance on the Visual Genome dataset with and without difficulty weighted training applied.}
\label{vg_test}
\end{table}

\section{Conclusion}
This paper investigated the use of class-level difficulty factors in multi-label classification problems, with methods developed to measure four distinct difficulty factors. The four proposed difficulty factors (frequency, visual variation, semantic abstraction and co-occurrence) are used to estimate class-level performance across datasets, with the most accurate regression performed when all four factors are combined. 

Difficulty-weighted CNN training is then investigated, with the best overall performance achieved when all four factors are summed to produce an overall difficulty weighting for each class. The proposed difficulty weighted approach to multi-label classification is shown to improve performance on both the WWW Crowd and Visual Genome datasets, with state-of-the-art performance observed on the WWW Crowd test set. Remarkably, even our baseline model, using only RGB features and test time augmentation, outperforms the existing state-of-the-art \cite{shao2016slicing} on WWW crowd by a substantial margin.
 This type of framework allows for additional difficulty factors or refined methods for computing existing factors to be included over time, allowing for more accurate performance prediction and difficulty weighted training.

Future work in this area can involve the application of difficulty factors to the development of a curriculum learning approach to multi-label classification, where the model is trained on easier classes first before more difficult classes are introduced to the model later on.


\subsection*{Acknowledgements}
This publication has emanated from research conducted with the financial support of Science Foundation Ireland (SFI) under grant number SFI/15/SIRG/3283 and SFI/12/RC/2289. The authors would like to thank Huawei Technologies, Co., Ltd. and Huawei Ireland Research Center in Dublin, Ireland who kindly provided funding and assistance in this project, filed under code YBN2017120005.

\bibliographystyle{IEEEbib}
\bibliography{icme2020template}

\end{document}